\documentclass[journal]{IEEEtran}

\usepackage[noadjust]{cite}
\usepackage{booktabs}
\usepackage[utf8]{inputenc}
\usepackage{graphicx}
\usepackage{acronym}
\usepackage{authblk}
\usepackage{blindtext}
\usepackage{lettrine}
\usepackage{adjustbox}
\usepackage{amsmath}
\usepackage{cases}
\usepackage{xcolor}
\usepackage{cleveref}
\usepackage{pifont}
\usepackage{url}
\ifCLASSOPTIONcompsoc
    \usepackage[caption=false, font=normalsize, labelfont=sf, textfont=sf]{subfig}
\else
\usepackage[caption=false, font=footnotesize]{subfig}
\fi
\usepackage{tabu}
\usepackage{tikz}

\usepackage{multirow}
\usepackage{threeparttable}
\usepackage{color, colortbl}
\definecolor{Gray}{gray}{0.9}

\acrodef{DL}[DL]{Deep Learning}
\acrodef{STE}[STE]{Straight-Through Estimator}
\acrodef{NN}[NN]{Neural Network}
\acrodef{SNN}[SNN]{Spiking Neural Network}
\acrodef{QNN}[QNN]{Quantized Neural Network}
\acrodef{LR}[LR]{Learning Rate}
\acrodef{QSNN}[QSNN]{Quantized SNN}
\acrodef{MSE}[MSE]{Mean Square Error}

\usepackage{cite}

\begin{document}
\title{Navigating Local Minima in Quantized Spiking Neural Networks}

\author{Jason~K.~Eshraghian*,
Corey~Lammie*,
Mostafa~Rahimi~Azghadi,
and~Wei~D.~Lu

\thanks{J. K. Eshraghian and W. D. Lu are with the Department of Electrical Engineering and Computer Science, University of Michigan, Ann Arbor, Michigan 48109, USA. J. K. Eshraghian is also with the Department of Computer Science and Software Engineering, The University of Western Australia, Crawley, WA 6009, Australia. (e-mail: \{jasonesh, wluee\}@umich.edu).}%
\thanks{C.~Lammie and M.~R.~Azghadi are with the College of Science and Engineering, James Cook University, Queensland 4814, Australia. (e-mail: \{corey.lammie, mostafa.rahimiazghadi\}@jcu.edu.au).}
\thanks{*J. K. Eshraghian and C. Lammie contributed equally to this manuscript.}}%
\maketitle

\begin{abstract}
Spiking and Quantized \acp{NN} are becoming exceedingly important for hyper-efficient implementations of \ac{DL} algorithms. However, these networks face challenges when trained using error backpropagation, due to the absence of gradient signals when applying hard thresholds. The broadly accepted trick to overcoming this is through the use of biased gradient estimators: surrogate gradients which approximate thresholding in \acp{SNN}, and \acp{STE}, which completely bypass thresholding in \acp{QNN}. While noisy gradient feedback has enabled reasonable performance on simple supervised learning tasks, it is thought that such noise increases the difficulty of finding optima in loss landscapes, especially during the later stages of optimization. By periodically boosting the \ac{LR} during training, we expect the network can navigate unexplored solution spaces that would otherwise be difficult to reach due to local minima, barriers, or flat surfaces. This paper presents a systematic evaluation of a cosine-annealed \ac{LR} schedule coupled with weight-independent adaptive moment estimation as applied to \acp{QSNN}.
We provide a rigorous empirical evaluation of this technique on high precision and 4-bit quantized \acp{SNN} across three datasets, demonstrating (close to) state-of-the-art performance on the more complex datasets. Our source code is available at this link: 
\url{https://github.com/jeshraghian/QSNNs}.

\end{abstract}

\IEEEpeerreviewmaketitle

\section{Introduction}
\lettrine{L}{ow-power} implementations of \acp{NN} are essential for operation on portable, edge devices~\cite{Azghadi2020, shi2016edge, yang2021adaptive}. %
Most resource-constrained algorithmic options either reduce memory usage or memory access frequency. As one example, 
\acp{QNN} compress the possible parameter space, such that weights are represented with limited resolution, reducing resource requirements~\cite{gong2014compressing, lee2021quantized, chen2015compressing, eshraghian2022fine}. %
As another example, \acp{SNN} take inspiration from biological neurons, which communicate via voltage spikes. Spikes may coarsely be treated as binary events, where a spike either occurs, or it does not \cite{maass1997networks}.
This is represented using binarized activations: an activation of `1' trades the multiply step with memory read-out, and `0' simply bypasses the need for memory access. This strict limitation is offset by representing information in the time domain rather than the binarized amplitude, i.e., \textit{when} the spike is triggered, rather than \textit{what} value the spike is~\cite{bohte2002error}.

\begin{figure}[!t]
\centering
\includegraphics[width=0.5\textwidth]{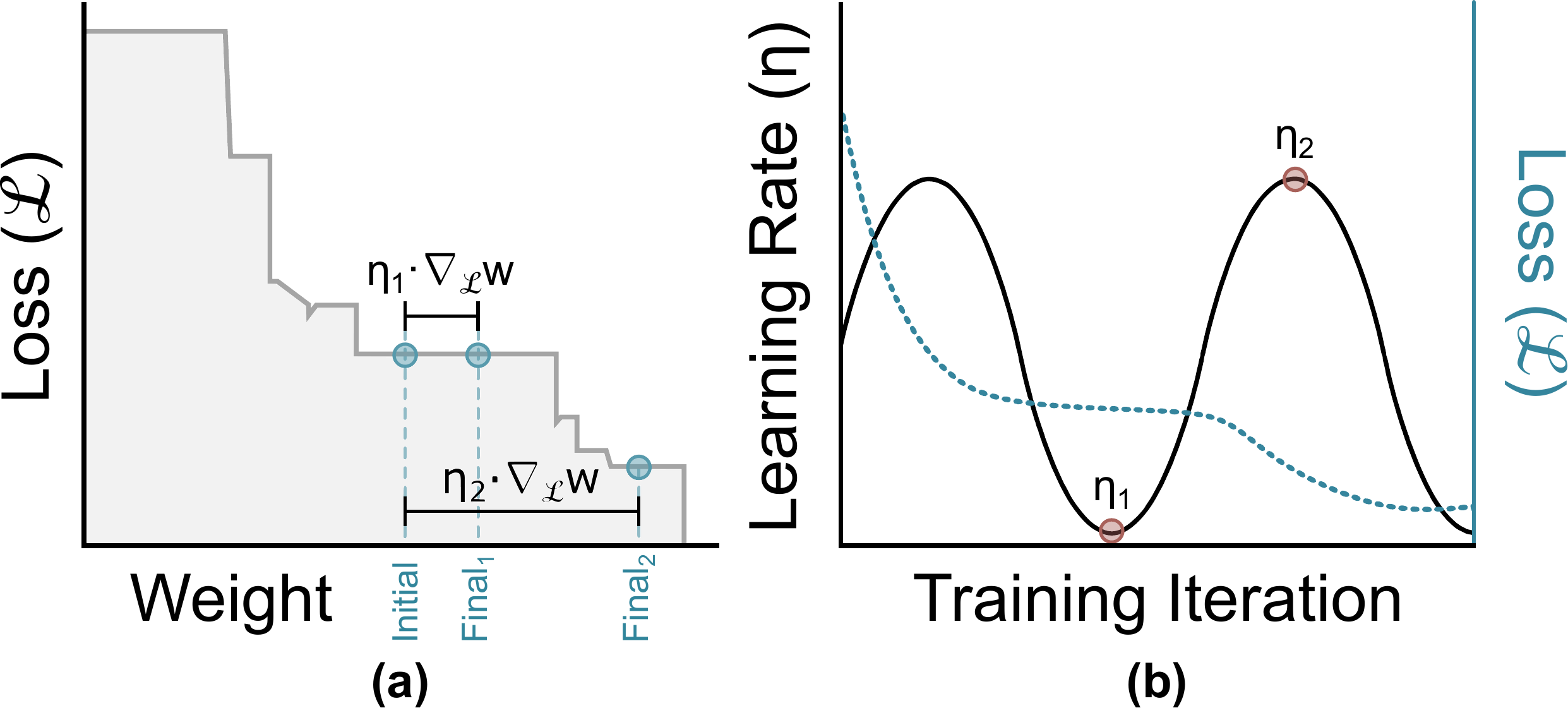}
\caption{Periodic scheduling can enable \acp{SNN} to overcome flat surfaces and local minima. (a) When the \ac{LR} is boosted during training using a cyclic scheduler, it is given another chance to reduce the loss with different initial conditions. (b) Loss evolution across training iterations. While the loss appears to converge, subsequent \ac{LR} boosting enables it to traverse more optimal solutions.}
\label{fig:overview}
\end{figure}

\acp{QSNN} offer orders of magnitude of power and latency improvement over full precision networks, but several approximations must be made during training before implementation using neuromorphic hardware~\cite{davies2018loihi, frenkel2019morphic, merolla2014million, eshraghian2022memristor, furber2014spinnaker, schmitt2017neuromorphic, neckar2018braindrop}. These approximations are applied during weight compression and gradient calculation to enable error backpropagation. While \acp{NN} are error tolerant to a degree, noise (e.g., quantization noise) often slows training convergence~\cite{bartunov2018assessing, lammie2021modeling}, or leads to suboptimal solutions if weights become stuck in absence of informative error signals~\cite{refinetti2021align, lammie2021memristive}\footnote{Non-systematic noise has been used to speed up convergence by overcoming saddle points/local minima \cite{rahimi2020complementary, cai2020power, kang2021build}, and also as a `pseudo-regularizer' to prevent overfitting \cite{chen2021analog}, though demonstrations are limited to simple tasks.}.  
\acp{QSNN} are especially susceptible to becoming stuck at 
suboptimal barriers and local minima during training, at excessive distance from the Bayes optimal error.
A large \ac{LR} leads to large weight changes at the risk of jumping over ideal solutions, but may also be required to traverse across flat loss surfaces in later stages of training (Fig.~\ref{fig:overview}(a)). 

When training \acp{QSNN}, we show the use of a periodic cosine-annealed \ac{LR} schedule~\cite{loshchilov2017sgdr} when coupled together with weight-independent adaptive decay~\cite{kingma2014adam}, as depicted in Fig.~\ref{fig:overview}, outperforms alternative, more commonly adopted schedules.
By periodically boosting the \ac{LR} at repeated intervals, this can be thought of as providing the network with another chance to search for a more optimal solution with potentially improved initial conditions. Although large learning rates may lead to divergence, early stopping with sufficient patience allows the network to revert back to a previous optimal state.
We provide a rigorous empirical analysis across three datasets ranging from simple to increasingly difficult: MNIST \cite{lecun1998gradient}, FashionMNIST \cite{xiao2017fashion}, and DVS128 Gesture \cite{amir2017low}. 

\noindent Our specific contributions are as follows:
\begin{enumerate}
    \item Our baseline \acp{SNN} perform competitively against previously reported results, without exotic techniques which increase memory and computational complexity;
    \item Our approach is tested on high precision \acp{SNN}, and 4-bit \acp{QSNN}, which are systematically compared against networks trained without scheduling, and with step decay, and loss-dependent schedules;
    \item Using annealed scheduling, we achieve the best performance for the FashionMNIST and DVS128 Gesture datasets when compared to other more commonly adopted schedules. Performance degradation when quantizing weights is also less to hyperparameters associated with non-periodic schedules.
\end{enumerate}

Furthermore, our empirical analysis across various datasets and hyperparameter sweeps provides insight into how \acp{QSNN} react to different optimization conditions and noise sources (e.g., dropout, surrogate gradients).

\section{Background and Related Work}\label{background}
\subsection{Spiking Neuron Model}
The spiking neuron model used is a single-state leaky integrate-and-fire neuron governed by the discrete-time dynamics:
\begin{equation}\label{eq:lif}
     u^j_{t+1} = \beta u^j_t + \sum_iw^{ij}z^i_t - z_t^j\theta,
\end{equation}
\begin{equation} \label{eq:spk}
    z^j_{t} =
    \begin{cases}
      1,  & \text{\rm if $u^j_t >\theta$} \\ 
      0, & \text{otherwise}, \\
    \end{cases}  
\end{equation}

\noindent which is derived using the forward Euler method to solve the continuous-time representation~\cite{eshraghian2021training}. Here, $u^j_{t}$ is the hidden state (membrane potential) of neuron $j$ at time $t$; $\beta$ is the membrane potential decay rate; $w^{ij}$ is the synaptic weight between neurons $i$ and $j$; and the final term of (\ref{eq:lif}) resets the state by subtracting the threshold $\theta$ each time an output spike $z^j_t \in \{0,1\}$ is generated.

\subsection{Hard Thresholds in \acp{QSNN}}
The sources of approximation in calculating the gradients of \acp{QNN} and \acp{SNN} are quite similar~\cite{lu2020exploring}. 
\acp{SNN} apply a hard threshold to a neuron's hidden state to generate a spike (\ref{eq:spk}), and \acp{QNN} apply a threshold to full precision weights to obtain fixed precision integers. Hard thresholds are non-differentiable, which null out gradient signals. Consequently, current best practices employ approximations that calculate biased gradient estimators during the backward pass. To solve the non-differentiability of \acp{SNN}, training \acp{SNN} broadly use surrogate gradients that `smooth' the thresholding function during training \cite{o2013real, neftci2019surrogate, hunsberger2015spiking}. Addressing the non-differentiability of quantization uses \acp{STE}, which simply bypasses the threshold operator in quantization-aware training schemes~\cite{hubara2016binarized, bengio2013estimating, lee2016training, o2016deep, hinton2012}. High precision weights are used to accumulate surrogate (approximate) gradients in backpropagation, while the quantized weights are used in the forward-pass. To formalize, let $w_r$ be the full precision proxy for the quantized weight $w_q$: 
\begin{equation}\label{eq:ste}
    \frac{\partial w_q}{\partial w_r} = 1.
\end{equation}
 
\noindent However, an infinitesimally small change of a high precision weight will not always change the quantized weight, nor will it always trigger a change of spiking activity\footnote{One caveat is that spike timing may be modified for certain neuron models, e.g., spike response models~\cite{bohte2002error}. We focus here on spiking neuron models with discontinuous state-changes.}. There is a risk that the effect of weight updates are absorbed into the subthreshold dynamics of spiking neurons, which causes the loss at the output to become identical across training iterations. Put succinctly, \acp{QSNN} are effectively more challenging to train than conventional \acp{NN}.

\subsection{Mitigating the Impact Of Quantization in \acp{QNN}}
Existing mitigation techniques require additional memory or increased computational complexity during training~\cite{putra2021}. Specific methods include taking the loss as a function of a continuous variable rather than single-bit spikes, such as spike time~ \cite{kheradpisheh2021bs4nn}, or membrane potential~\cite{schaefer2020quantizing}. Such methods hold much potential when the cost of feedforward inference outweighs the importance of the training step.

In our experiments, we avoid the modification of neuron models, or any other technique that is not common practice in \ac{DL}, to avoid adding overhead at runtime. Specifically, we constrain our analysis to only make modifications to the evolution of the \ac{LR} across training iterations together with an extensive hyperparameter search.

\subsection{Periodic \ac{LR} Schedules}
The most broadly used strategy for adjusting the \ac{LR} is exponential decay~\cite{he2016deep}. 
 For example, He \textit{et al.} apply a step decay at a rate of 0.1/30 epochs. 
Recently, cosine annealing has shown superior performance where the \ac{LR} follows a half cosine curve~\cite{loshchilov2017sgdr, he2019bag}. \ac{LR} restarts can encourage the model to move from one local minimum to another, with the expectation that it allows the model to explore uncharted regions of the solution space, and therefore lead to a more accurate result.
This may be especially important in networks that are susceptible to becoming stuck in sub-optimal regions, such as \acp{QSNN}. 
We compute the \ac{LR}, $\eta_t$, in our experiments using the following scheduler inspired by~\cite{loshchilov2017sgdr, he2019bag}:
\begin{equation}
    \eta_t = \frac{1}{2} \eta \Big(1 + \text{cos}\Big(\frac{\pi t}{T} \Big) \Big),
\end{equation}

\noindent where $\eta$ is the initial \ac{LR}, $t$ is the iteration, and $T$ is the period of the schedule. Several prior works have used cosine-based scheduling to train \acp{SNN} via error backpropagation, where its use is typically incidental to proposing alternative novel methods to train networks. For example, Refs. \cite{cordone2021learning} and \cite{liu2021sstdp} apply periodic scheduling to strided sparse convolutional \acp{SNN} and spike-time based learning rules, respectively, both to highly competitive results. We aim to decouple the effect of emerging training algorithms from that of best deep learning practices. The work in \cite{shen2021backpropagation} uses a half cosine schedule without periodicity, where it is expected the absence of discontinuities in the \ac{LR} improves network performance \cite{loshchilov2017sgdr}.

\section{Methods}\label{methods}
\noindent \textbf{Testing:} Our hypothesis is that cosine scheduling can improve performance by escaping local minima in challenging networks with ill-defined gradients, such as \acp{SNN} and \acp{QSNN}. We assess performance on three datasets using \acp{SNN} and \acp{QSNN}. We use uniform 4-bit weight quantization, as it is closely compatible with several lightweight academic neuromorphic chips~\cite{frenkel2019morphic, schmitt2017neuromorphic, lee2020neuromorphic}, and also poses a greater challenge to train over more conventional 8-bit integer weights~\cite{davies2018loihi}. 

\noindent \textbf{Gradient Approximations:} Quantization-aware training is used for the 4-bit \acp{QSNN}, and a threshold-shifted fast sigmoid surrogate gradient is applied to the spike non-differentiability:
\begin{equation}\label{eq:dfs}
    \frac{\partial \tilde{z}}{\partial u}=\frac{1}{(1+k|\theta - u|)^2}.
\end{equation}

\noindent $k$ modifies the slope of the surrogate gradient and is tuned during a hyperparameter search. A smaller value of $k$ corresponds to a larger approximation.

\begin{figure}[!t]
\centering
\includegraphics[width=0.45\textwidth]{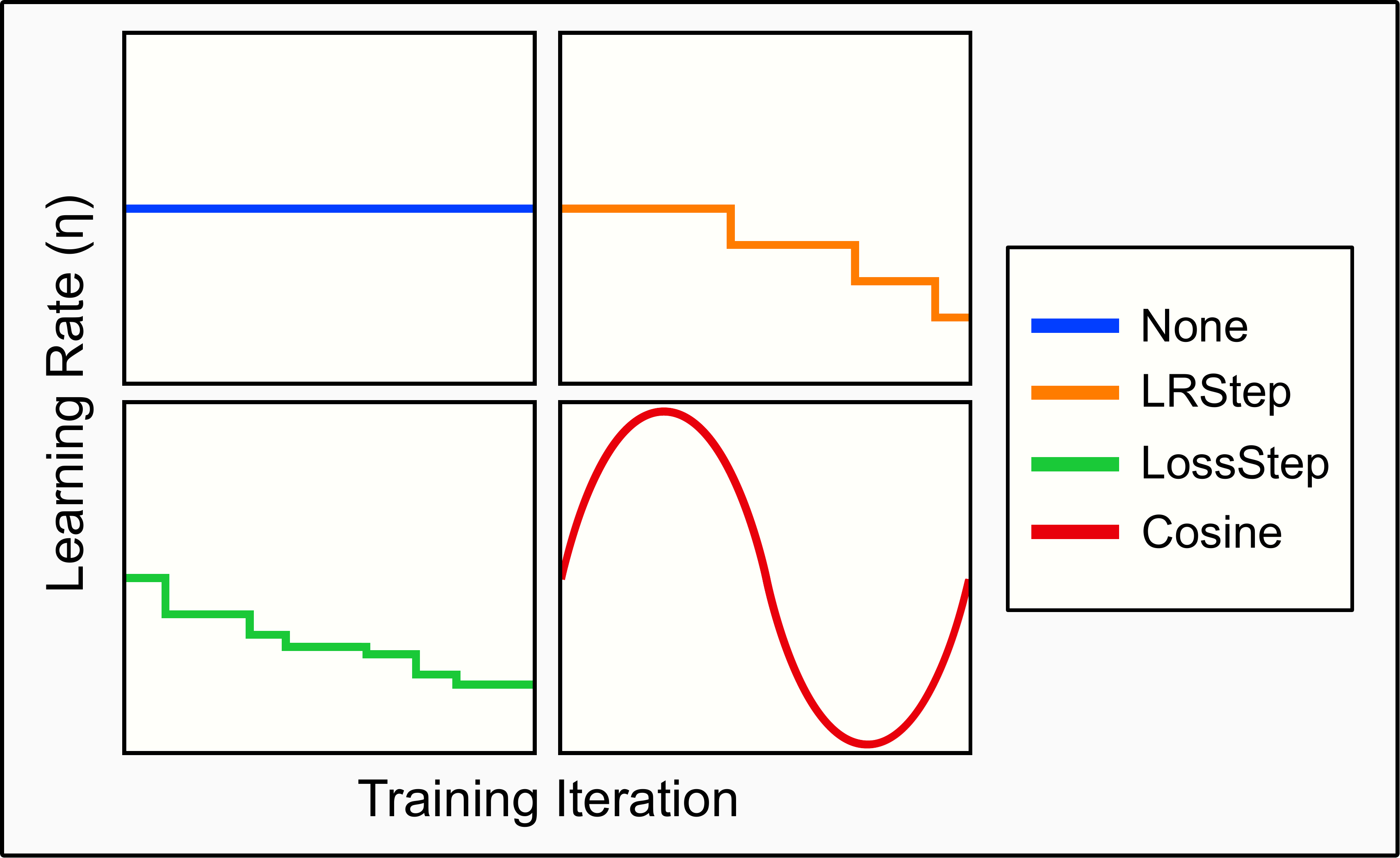}
\caption{Various learning rate schedules which were evaluated. These schedules are depicted for illustration purposes, and are not representative of the exact schedules used in Section~\ref{experimental}. For LRStep, the step interval and step height can be configured arbitrarily. For LossStep, the iteration and loss thresholds, in addition to the step height can be configured arbitrarily. For Cosine, the period and amplitude can be configured arbitrarily.}
\label{fig:sched}
\end{figure}

\noindent \textbf{Optimization:} Each network is tested without scheduling, step decay (`LRStep'), loss-dependent decay (`LossStep'), and cosine annealing, all of which are depicted in \Cref{fig:sched}. The schedule is updated for each minibatch. The Adam optimizer has shown empirically good performance on recurrent and spiking networks, and our early experiments showed it outperformed stochastic gradient descent. Consequently, we combine Adam's weight-adaptive learning rate \cite{kingma2014adam} with cosine annealing \cite{loshchilov2017sgdr} with a period of 10 epochs, repeated over the entire range of epochs. Further detail pertaining to network initialization, and network and neuron hyperparameters is provided in~\url{https://github.com/jeshraghian/QSNNs}.

\begin{table*}[!ht]\centering \caption{\ac{QSNN} Parameters.}\label{tab:params}
\begin{threeparttable}
\begin{tabular}{ccccccccccc} \toprule \toprule
\textbf{Dataset} & \textbf{Precision} & \textbf{Batch Size} & \textbf{Decay Rate $\beta$} & \textbf{Threshold $\theta$} & \textbf{Slope $k$} & \textbf{Initial LR $\eta$} & \textbf{Grad. Clip} & \textbf{Weight Clip} & \textbf{BatchNorm} & \textbf{Dropout} \\ \midrule
MNIST & flt32 & 128 & 0.92 & 2.0 & 6.0 & 1.9e-3 & \ding{51} & \ding{51} & \ding{51} & 0.09\\
MNIST & int4 & 128 & 0.99 & 2.9 & 13.8 & 5.4e-3 & \ding{51} & \ding{55} & \ding{51} & 0.00 \\ \midrule
FMNIST & flt32 & 128 & 0.39 & 1.5 & 7.7 & 2.0e-3 & \ding{51} & \ding{51} & \ding{51} & 0.13 \\
FMNIST & int4 & 128 & 0.97 & 2.5 & 5.6 & 2.9e-3 & \ding{55} & \ding{55} & \ding{51} & 0.07 \\ \midrule
DVS128 & flt32 & 16 & 0.72 & 2.5 & 9.7 & 2.4e-3& \ding{55} & \ding{51} & \ding{51} & 0.29  \\
DVS128 & int4 & 16 & 0.61 & 0.4 & 4.4 & 2.6e-3 & \ding{51} & \ding{51} & \ding{55} & 0.20  \\
\bottomrule \bottomrule

\end{tabular}
 \begin{tablenotes}
    \item[1] Optimal parameters selected from 500 separate trials of 5 epochs each using a tree-structured Parzen Estimator algorithm to randomly sample from the wide hyperparameter space performed using Optuna \cite{akiba2019optuna, bergstra2011algorithms}. The above parameters were used for full training runs until early stopping terminated the process.
  \end{tablenotes}
 \end{threeparttable}
\end{table*}

\noindent \textbf{Loss:}
To avoid offsetting the hardware benefits gained from operating on discrete variables, and to further `stress' the network under a discontinuous loss landscape mired with flat surfaces, we test our network under the more challenging constraint of using target spike counts for each class. The spikes of each output neuron $z^j_t$ are accumulated over time. The \ac{MSE} from the target spike count $c^j$ is measured, and then summed across $N$ output classes:
\begin{equation}\label{eq:loss}
    \mathcal{L}_{MSE} = \sum_j^{N}\sum_t(c^j - z^j_t)^2.
\end{equation}

\noindent The target in all cases adopts a similar approach to Shrestha and Orchard \cite{shrestha2018slayer}: the correct class $c$ is targeted to fire 80\% of all time steps, while incorrect classes are set to 20\% (intended to avoid the suppression of neuron activity). 
The predicted class is the neuron with the highest spike count in each simulation. 

\noindent \textbf{Architecture:} Each model uses 2 convolutional layers terminated with a dense layer. Each convolution operation was followed by average-pooling applied to the membrane potential rather than spikes, as pooling sparse outputs with many zero elements would arbitrarily drive the influence of spikes lower. The exact architecture for each dataset is provided as a footnote in Table~\ref{table:main_result}.

While learnable and heterogeneous neuronal parameters can be used to enhance network performance \cite{perez2021neural}, we avoid them here because individualising neuron parameters causes forward pass memory to scale with $\mathcal{O}(n)$, and during training with $\mathcal{O}(nT)$ where $n$ is the number of neurons and $T$ is the number of time steps. This otherwise counters benefits derived from quantization.

\noindent \textbf{Initialization: } Dense layers were initialized by uniformly sampling from \textit{U}($-\sqrt{a}$, $\sqrt{a}$):

\begin{equation}\label{eq:initlin}
    a = \frac{1}{N_{\rm in}}, 
\end{equation}

\noindent where $N_{\rm in}$ is the number of input features. Parameters in convolutional layers were also uniformly sampled:

\begin{equation}
    a = \frac{1}{C_{\rm in}N_x N_y},
\end{equation}

\noindent where $C_{\rm in}$ is the number of input channels, and $N_x$ and $N_y$ are the kernel dimensions. 

\newcolumntype{g}{>{\columncolor{Gray}}c}
\begin{table*}[!ht]
\centering
\caption{Test set accuracies for the MNIST, FashionMNIST, and DVS128 Gesture Datasets.}\label{table:main_result}
\begin{threeparttable}
\begin{tabular}{llgggcccggg} \toprule \toprule
\multirow{2}{*}{\textbf{Precision}} & \multirow{2}{*}{\textbf{Scheduler}} & \multicolumn{3}{c}{\textbf{MNIST}} & \multicolumn{3}{c}{\textbf{FashionMNIST}} & \multicolumn{3}{c}{\textbf{DVS128 Gesture}} \\
 &  & \multicolumn{1}{l}{\textbf{Best}} & \multicolumn{1}{l}{\textbf{Avg. ($n=3$)}} & \multicolumn{1}{l}{\textbf{$\sigma$}} & \multicolumn{1}{l}{\textbf{Best}} & \multicolumn{1}{l}{\textbf{Avg. ($n=3$)}} & \multicolumn{1}{l}{\textbf{$\sigma$}} & \multicolumn{1}{l}{\textbf{Best}} & \multicolumn{1}{l}{\textbf{Avg. ($n=3$)}} & \multicolumn{1}{l}{\textbf{$\sigma$}} \\
 \midrule
flt32 & None & \textbf{99.45} & \textbf{99.31} & 0.12 & 90.49 & 90.37 & 0.18 & 91.67 & 90.39 & 1.12 \\
flt32 & LRStep & 99.36 & 99.27 & 0.08 & 90.85 & 90.69 & 0.14 & 92.36 & 90.74 & 1.75 \\
flt32 & LossStep & 99.29 & 99.23 & 0.04 & 90.86 & 90.71 & 0.18 & 90.97 & 90.74 & 0.16 \\
flt32 & Cosine & 99.29 & 99.24 & 0.04 & \textbf{91.13} & \textbf{91.03} & 0.08 & \textbf{93.05} & \textbf{92.87} & 0.32 \\ \midrule
int4 & None & 99.27 & 99.23 & 0.03 & 90.35 & 90.29 & 0.07 & 90.27 & 89.58 & 0.69 \\
int4 & LRStep & 99.35 & 99.3 & 0.08 & 90.43 & 90.26 & 0.15 & 91.32 & 90.62 & 1.20 \\
int4 & LossStep & \textbf{99.36} & \textbf{99.31} & 0.05 & \textbf{90.93} & \textbf{90.87} & 0.07 & 91.32 & 88.77 & 2.13 \\
int4 & Cosine & 99.33 & \textbf{99.31} & 0.02 & \textbf{90.93} & 90.83 & 0.09 & \textbf{92.01} & \textbf{91.44} & 0.52 \\ \bottomrule \bottomrule
\end{tabular}
\begin{tablenotes}
    \item[1] MNIST Architecture: 16Conv5-AP2-64Conv5-AP2-1024Dense10. FashionMNIST Architecture: 16Conv5-AP2-64Conv5-AP2-1024Dense10. DVS128 Gesture Architecture: 16Conv5-AP2-32Conv5-AP2-8800Dense11. $^2$ $\sigma$: Sample standard deviation.
\end{tablenotes}
\end{threeparttable}
\end{table*}

\section{Experimental Results}\label{experimental}
\begin{figure*}[!t]
    \centering
    \includegraphics[width=0.95\textwidth]{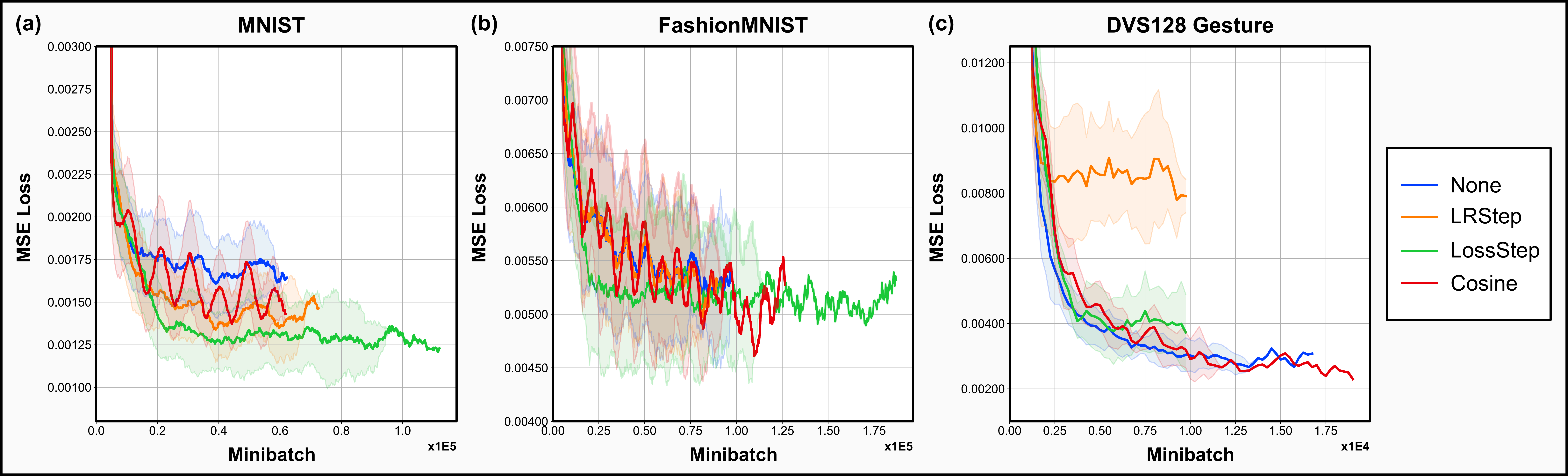}
    \caption{The moving average of the \ac{MSE} loss evolution for different schedules when training our \acp{QSNN} using the MNIST, FashionMNIST, and DVS128 Gesture datasets. As some \ac{LR} parameters were subject to hyperparameter optimization, they have not been visualized. LRStep terminates significantly earlier than other schedules as the learning rate drops faster without picking back up as with Cosine.}
    \label{fig:main_result}
\end{figure*}

Each experiment was run on a 16GB NVIDIA V100 GPU. Brevitas 0.7.1 was used to uniformly quantize network parameters~\cite{brevitas}, snnTorch 0.4.11 to construct spiking neuron models~\cite{eshraghian2021training}, PyTorch 1.10.1 for training~\cite{paszke2019pytorch}, in Python 3.7.9 with results recorded across three trials.
Hyperparameters for each dataset and precision combination are provided in \Cref{tab:params}. For all datasets, no data augmentation was used, the test set accuracy across trials is shown in~\Cref{table:main_result}, and the moving average of the \ac{MSE} loss during training of \acp{QNN} is shown in Fig.~\ref{fig:main_result}. For all implementations, `LRStep' halves the \ac{LR} every 15 epochs, `LossStep' halves the \ac{LR} each time the training loss does not improve, and `Cosine' is our application of annealing, which cycles every 10 epochs.

\subsection{MNIST}
The raw MNIST dataset was repeatedly passed to the network for 100 time steps of simulation, without encoding~\cite{lecun1998gradient}.
The MNIST dataset includes 60,000 28$\times$28 greyscale images of handwritten digits in the training set, and 10,000 in the test set~\cite{lecun1998gradient}.
Averaging across all 12 high precision trials ($4\times n$ trials), it took $\approx$65 epochs for early stopping to terminate training, which provided ample opportunity for scheduling to impact the learning process. 

For high precision weights, our proposed approach marginally underperforms when compared to alternative scheduling techniques, though other schedules cause high variance across trials.
However, the MNIST dataset is commonly regarded as too simple of a task, and so difficulty is marginally increased by using 4-bit quantized weights. Several broad observations are made.
Firstly, the experiments without scheduling drops from best to worst performance when going from high precision to fixed precision integers. This may indicate the use of scheduling becomes increasingly important as task complexity increases. Secondly, cosine annealing shows the best average accuracy for 4-bit weights and smallest variance (and therefore, consistency). Finally, the average performance with cosine annealing \textit{improved} when constraining weight precision (99.24\% to 99.31\%) - an unexpected result that is explored further in the next section. More complex datasets can offer more insight.

\subsection{FashionMNIST}
The FashionMNIST dataset is intended to be a more challenging `drop-in' replacement for MNIST, with identical resolution and dataset size, consisting of 10 classes of clothing items and accessories~\cite{xiao2017fashion}. The raw input was again passed to the network for 100 time steps of simulation without encoding.
We now find that cosine annealing outperforms across the board, and the lack of scheduling reduces accuracy significantly, as distinct from high precision MNIST. Quantization now has a slight detrimental impact on performance (degrading the average from 91.03\% to 90.83\% for cosine annealing), but the effect is observed to be quite small as \acp{QSNN} amortize the cost by accumulating spikes in continuous spaces (i.e., time and hidden state).

\subsection{DVS128 Gesture}
The DVS128 Gesture dataset is an event-based dataset with 11 different gesture classes, such as clapping, air guitar, etc., featuring temporal dynamics~\cite{amir2017low}.
The dataset is filmed with a dynamic vision sensor, which only processes sufficient changes in luminance, of resolution $128 \times 128$ with two channels: one for on-spikes (increases in luminance) and another for off-spikes (decreases in luminance) \cite{patrick2008128x}. 
To account for GPU memory constraints, spatial downsampling was applied ($32 \times 32$), events were integrated over a temporal resolution of 5~ms per input at a given sequence step, such that training samples were fit to a duration of 1~s each, which corresponds to a sequence length of 100 time steps.
Cosine annealing again outperforms the average test set accuracy of all other techniques, this time by a more significant margin (92.87\% for Cosine annealing, 90.74\% for the next best). This is accompanied by a considerably low standard deviation ($\sigma=0.32$). The variance is the smallest when weights are quantized ($\sigma=0.52$; next best: $\sigma=0.69$). Quantization has a more significant impact with increasing data complexity (reduction by $\sim$1.4\% for cosine annealing, while other schedules suffer greater degradation), but this still manages to outperform all other learning rate schedules in the full precision experiment. We note that most top-performing results on the DVS128 Gesture dataset are in excess of 93\% \cite{shrestha2018slayer}, though the model we use is reasonably smaller in comparison with only a marginal reduction in performance.

\section{Discussion}\label{sec:discussion}
\textbf{Reduced variance of accuracy: }
Cosine scheduling almost always achieved smaller variance of accuracy when compared to the alternatives. Periodically boosting the learning rate enables the network to explore alternative solution spaces, where if performance deviates too far from the ideal, early stopping with patience allows the network to revert back to the optimal solution. The low variance (across trials) coupled with the relatively low bias of cosine annealing may indicate that alternative schedules occasionally enable convergence to an optimal solution, but with less confidence than for cosine annealing. 

\textbf{Inconsistent effects of quantization noise: } 
Oddly, quantization noise enabled the network trained on the MNIST dataset using cosine scheduling to consistently \textit{improve}. While the same improvement from quantization does not hold for our experiments on the FashionMNIST and DVS128 Gesture datasets, it shows that \acp{SNN} are quite robust to quantization. We believe it is likely that QSNNs absorb a significant degree of the quantization noise into: i) the subthreshold dynamics of neurons \cite{sharmin2019comprehensive}, and ii) the high precision state-space of the spiking neuron. In the case of a task as simple as MNIST classification, the noise that does alter network activity may actually serve to prevent overfitting. We note that our analysis is fixed to weight quantization rather than state quantization. The dominant cost in neural networks is memory access and data transfer, and a neuron's state is a continuously changing variable that is more likely to be stored in high-speed cache rather than in main memory. For additional empirical analyses on quantized state variables, we recommend referring to the work in \cite{schaefer2020quantizing, yousefzadeh2019conversion}.

\textbf{Alternative noise sources are reduced in quantized networks: } 
When moving from high precision to fixed precision networks, the dropout rate in dense layers consistently decreases (MNIST: $9\% \rightarrow 0\%$, FashionMNIST: $13\% \rightarrow 7\%$, DVS128 Gesture: $29\% \rightarrow 20\%$). We offer two possible reasons that support why lower dropout rates are preferred in QSNNs. Firstly, this may be the network's attempt to offset quantization noise by reducing other noise sources in the network (i.e., dropout probability). Secondly, the effect of dropout is that a full network (or layer) is treated as an ensemble of various sub-networks that share parameters at training time. This reduces the number of pathways available for backpropagation to be applied through the unrolled computational graph of the network. While this may not be as significant an issue in non-spiking networks, high rates of dropout lead to a smaller number of neuronal connections, which may lead to weaker gradient signals. This is further exacerbated by the stringent training conditions we applied, i.e. backpropagating through discrete spike counts rather than continuous membrane potentials. Ultimately, reducing the dropout rate enables the output layer to have higher fan-in, which increases the likelihood of early layer spiking activity propagating all the way through, and triggering a change to the final loss.

\textbf{Boosted gradient signals:}
When quantizing weights, the preferred surrogate gradient slope $k$ decreased by 27\% and 54\% for FashionMNIST and DVS128 Gesture, respectively. The surrogate gradient is limited to the range of $\partial \tilde{z} / \partial u \in [0, 1]$, which can lead to vanishing gradients when training deep networks.\footnote{For the same reasons why the deep learning community shifted from sigmoid activations to ReLU \cite{he2016deep}.} With respect to \Cref{eq:dfs}, as $k \rightarrow 0$, the fast sigmoid surrogate function approaches a straight line with a constant gradient $\partial \tilde{z}/\partial u \rightarrow 1$. That is to say, in the limit, the surrogate gradient converges to a \ac{STE}. A smaller value of $k$ promotes gradient signal propagation by reducing the risk of vanishing gradients, which is increasingly important for quantized networks that are susceptible to flat loss landscapes.

\textbf{The trade-off between long training runs and optimal solutions:} While weight quantization increased the difficulty of reaching an optimal result, cosine annealing reduced the impact significantly by persistently searching new solution spaces. As can be seen in Fig.~\ref{fig:main_result}, early stopping was applied to all other schedules considerably earlier when trained on the FashionMNIST and DVS128 Gesture datasets. For example, the best high precision trial using cosine annealing with the DVS128 Gesture data set (93.05\%, \Cref{table:main_result}) is not too far from the next best performing approach (92.36\%, LRStep) which suggests that both are capable of finding optimal solutions. But the large variation and low average for LRStep shows that it is also susceptible to being stuck in suboptimal regions, unable to escape in absence of periodic LR boosts. The extended training duration for cosine annealing is a cost to consider when relying on early stopping coupled with periodic schedules, though we note that all other schedules were afforded the same early stopping criterion.

\textbf{Quantization increases the difficulty of finding optimal solutions, and periodic scheduling can offset this difficulty: } This observation becomes most clear when comparing the high precision performance of DVS128 Gesture to the fixed precision performance in \Cref{table:main_result}. The top performing trial for all schedules are within a range of 1.44\% in the high precision case, which expands to 1.86\% in the quantized network due to substantial performance degradation of non-periodic schedules. This may indicate that quantization does not completely eliminate optimal solutions, but rather makes them harder to find. We also note there may be more optimal step sizes and frequencies to improve results obtained from the alternative schedules in the quantized case (i.e., LRStep and LossStep). As examples, LRStep underperformed with a schedule rate of 15 epochs (average accuracy of 86.11\%), and had to be increased to 20 epochs (average accuracy of 91.32\%). The frequency of LossStep also had to be reduced, where the LR was halved only when the loss worsened for two consecutive epochs. This may potentially be due to the smaller size of the DVS128 Gesture dataset when compared to MNIST and FashionMNIST, though the period of cosine annealing was completely robust and did not require any modification to maintain good performance. The alternative schedules therefore add an additional free (and highly sensitive) hyperparameter which periodic scheduling shows robustness against.

\section{Conclusion}\label{sec:conclusion}
We have demonstrated the ability of periodic scheduling as a promising method to continue finding optimal solutions in cases where other schedules cease to improve network performance. Performance degradation from weight quantization impacts non-periodic schedules substantially more than cosine-annealing, and has an increasing impact with data complexity. We have also shown that \acp{SNN} can be made to be robust to weight quantization which offers hardware savings across both memory and computation, given sufficient exploration of the solution space which is more likely to be achieved using periodic schedules. The online repository containing high precision and quantized \ac{SNN} experiments across the reported datasets can be accessed at this link: 
\url{https://github.com/jeshraghian/QSNNs}.

\color{black}
\bibliographystyle{IEEEtran}
\bibliography{References}

\begin{thebibliography}{10}
\providecommand{\url}[1]{#1}
\csname url@samestyle\endcsname
\providecommand{\newblock}{\relax}
\providecommand{\bibinfo}[2]{#2}
\providecommand{\BIBentrySTDinterwordspacing}{\spaceskip=0pt\relax}
\providecommand{\BIBentryALTinterwordstretchfactor}{4}
\providecommand{\BIBentryALTinterwordspacing}{\spaceskip=\fontdimen2\font plus
\BIBentryALTinterwordstretchfactor\fontdimen3\font minus
  \fontdimen4\font\relax}
\providecommand{\BIBforeignlanguage}[2]{{%
\expandafter\ifx\csname l@#1\endcsname\relax
\typeout{** WARNING: IEEEtran.bst: No hyphenation pattern has been}%
\typeout{** loaded for the language `#1'. Using the pattern for}%
\typeout{** the default language instead.}%
\else
\language=\csname l@#1\endcsname
\fi
#2}}
\providecommand{\BIBdecl}{\relax}
\BIBdecl

\bibitem{Azghadi2020}
M.~R. Azghadi \emph{et~al.}, ``{Hardware Implementation of Deep Network
  Accelerators Towards Healthcare and Biomedical Applications},'' \emph{IEEE
  Trans. on Biomed. Circuits Syst.}, vol.~14, no.~6, pp. 1138--1159, 2020.

\bibitem{shi2016edge}
W.~Shi, J.~Cao, Q.~Zhang, Y.~Li, and L.~Xu, ``Edge computing: Vision and
  challenges,'' \emph{IEEE Internet of Things J.}, vol.~3, no.~5, pp. 637--646,
  2016.

\bibitem{yang2021adaptive}
Y.~Yang, N.~D. Truong, J.~K. Eshraghian, A.~Nikpour, and O.~Kavehei,
  ``Adaptive, unlabeled and real-time approximate-learning platform (aura) for
  personalized epileptic seizure forecasting,'' \emph{medRxiv}, 2021.

\bibitem{gong2014compressing}
Y.~Gong, L.~Liu, M.~Yang, and L.~Bourdev, ``Compressing deep convolutional
  networks using vector quantization,'' \emph{arXiv preprint arXiv:1412.6115},
  2014.

\bibitem{lee2021quantized}
J.~Lee, J.~K. Eshraghian, S.~Kim, K.~Eshraghian, and K.~Cho, ``Quantized
  convolutional neural network implementation on a parallel-connected memristor
  crossbar array for edge ai platforms,'' \emph{Journal of Nanoscience and
  Nanotechnology}, vol.~21, no.~3, pp. 1854--1861, 2021.

\bibitem{chen2015compressing}
W.~Chen, J.~Wilson, S.~Tyree, K.~Weinberger, and Y.~Chen, ``Compressing neural
  networks with the hashing trick,'' in \emph{Intl. Conf. on machine
  learning}.\hskip 1em plus 0.5em minus 0.4em\relax PMLR, 2015, pp. 2285--2294.

\bibitem{eshraghian2022fine}
J.~K. Eshraghian and W.~D. Lu, ``The fine line between dead neurons and
  sparsity in binarized spiking neural networks,'' \emph{arXiv preprint
  arXiv:2201.11915}, 2022.

\bibitem{maass1997networks}
W.~Maass, ``Networks of spiking neurons: the third generation of neural network
  models,'' \emph{Neural networks}, vol.~10, no.~9, pp. 1659--1671, 1997.

\bibitem{bohte2002error}
S.~M. Bohte, J.~N. Kok, and H.~La~Poutre, ``Error-backpropagation in temporally
  encoded networks of spiking neurons,'' \emph{Neurocomputing}, vol.~48, no.
  1-4, pp. 17--37, 2002.

\bibitem{davies2018loihi}
M.~Davies \emph{et~al.}, ``Loihi: A neuromorphic manycore processor with
  on-chip learning,'' \emph{IEEE Micro}, vol.~38, no.~1, pp. 82--99, 2018.

\bibitem{frenkel2019morphic}
C.~Frenkel, J.-D. Legat, and D.~Bol, ``Morph{IC}: A 65-nm 738k-synapse/mm$^{2}$
  quad-core binary-weight digital neuromorphic processor with stochastic
  spike-driven online learning,'' \emph{IEEE Trans. Biomed. Circuits Syst.},
  vol.~13, no.~5, pp. 999--1010, 2019.

\bibitem{merolla2014million}
P.~A. Merolla, J.~V. Arthur, R.~Alvarez-Icaza, A.~S. Cassidy, J.~Sawada,
  F.~Akopyan, B.~L. Jackson, N.~Imam, C.~Guo, Y.~Nakamura \emph{et~al.}, ``A
  million spiking-neuron integrated circuit with a scalable communication
  network and interface,'' \emph{Science}, vol. 345, no. 6197, pp. 668--673,
  2014.

\bibitem{eshraghian2022memristor}
J.~K. Eshraghian, X.~Wang, and W.~D. Lu, ``Memristor-based binarized spiking
  neural networks: Challenges and applications.'' \emph{IEEE Nanotechnology
  Magazine}, 2022.

\bibitem{furber2014spinnaker}
S.~B. Furber, F.~Galluppi, S.~Temple, and L.~A. Plana, ``The {SpiNNaker}
  project,'' \emph{Proc. of the IEEE}, vol. 102, no.~5, pp. 652--665, 2014.

\bibitem{schmitt2017neuromorphic}
S.~Schmitt, J.~Kl{\"a}hn, G.~Bellec, A.~Gr{\"u}bl, M.~Guettler, A.~Hartel,
  S.~Hartmann, D.~Husmann, K.~Husmann, S.~Jeltsch \emph{et~al.}, ``Neuromorphic
  hardware in the loop: Training a deep spiking network on the brainscales
  wafer-scale system,'' in \emph{2017 Intl. Joint Conf. on Neural Netw.
  (IJCNN)}.\hskip 1em plus 0.5em minus 0.4em\relax IEEE, 2017, pp. 2227--2234.

\bibitem{neckar2018braindrop}
A.~Neckar, S.~Fok, B.~V. Benjamin, T.~C. Stewart, N.~N. Oza, A.~R. Voelker,
  C.~Eliasmith, R.~Manohar, and K.~Boahen, ``Braindrop: A mixed-signal
  neuromorphic architecture with a dynamical systems-based programming model,''
  \emph{Proceedings of the IEEE}, vol. 107, no.~1, pp. 144--164, 2018.

\bibitem{bartunov2018assessing}
S.~Bartunov, A.~Santoro, B.~A. Richards, L.~Marris, G.~E. Hinton, and
  T.~Lillicrap, ``Assessing the scalability of biologically-motivated deep
  learning algorithms and architectures,'' \emph{32nd Conf. on Neural Inf.
  Process. Syst. (NeurIPS 2018)}, 2018.

\bibitem{lammie2021modeling}
C.~Lammie, W.~Xiang, and M.~R. Azghadi, ``Modeling and simulating in-memory
  memristive deep learning systems: An overview of current efforts,''
  \emph{Array}, p. 100116, 2021.

\bibitem{refinetti2021align}
M.~Refinetti, S.~d’Ascoli, R.~Ohana, and S.~Goldt, ``Align, then memorise:
  The dynamics of learning with feedback alignment,'' in \emph{Intl. Conf. on
  Machine Learning}.\hskip 1em plus 0.5em minus 0.4em\relax PMLR, 2021, pp.
  8925--8935.

\bibitem{lammie2021memristive}
C.~Lammie, J.~K. Eshraghian, W.~D. Lu, and M.~R. Azghadi, ``Memristive
  stochastic computing for deep learning parameter optimization,'' \emph{IEEE
  Trans. on Circuits and Syst. II: Exp. Briefs}, vol.~68, no.~5, pp.
  1650--1654, 2021.

\bibitem{rahimi2020complementary}
M.~Rahimi~Azghadi, Y.-C. Chen, J.~K. Eshraghian, J.~Chen, C.-Y. Lin,
  A.~Amirsoleimani, A.~Mehonic, A.~J. Kenyon, B.~Fowler, J.~C. Lee
  \emph{et~al.}, ``Complementary metal-oxide semiconductor and memristive
  hardware for neuromorphic computing,'' \emph{Advanced Intelligent Systems},
  vol.~2, no.~5, p. 1900189, 2020.

\bibitem{cai2020power}
F.~Cai, S.~Kumar, T.~Van~Vaerenbergh, X.~Sheng, R.~Liu, C.~Li, Z.~Liu,
  M.~Foltin, S.~Yu, Q.~Xia \emph{et~al.}, ``Power-efficient combinatorial
  optimization using intrinsic noise in memristor hopfield neural networks,''
  \emph{Nature Electronics}, vol.~3, no.~7, pp. 409--418, 2020.

\bibitem{kang2021build}
S.~M. Kang, D.~Choi, J.~K. Eshraghian, P.~Zhou, J.~Kim, B.-S. Kong, X.~Zhu,
  A.~S. Demirkol, A.~Ascoli, R.~Tetzlaff \emph{et~al.}, ``How to build a
  memristive integrate-and-fire model for spiking neuronal signal generation,''
  \emph{IEEE Transactions on Circuits and Systems I: Regular Papers}, vol.~68,
  no.~12, pp. 4837--4850, 2021.

\bibitem{chen2021analog}
Y.-C. Chen, J.~K. Eshraghian, I.~Shipley, and M.~Weiss, ``Analog synaptic
  behaviors in carbon-based self-selective rram for in-memory supervised
  learning,'' in \emph{2021 IEEE 71st Electronic Components and Technology
  Conference (ECTC)}.\hskip 1em plus 0.5em minus 0.4em\relax IEEE, 2021, pp.
  1645--1651.

\bibitem{loshchilov2017sgdr}
I.~Loshchilov and F.~Hutter, ``{SGDR}: Stochastic gradient descent with warm
  restarts,'' \emph{Intl. Conf. on Learning Representations}, 2017.

\bibitem{kingma2014adam}
D.~P. Kingma and J.~Ba, ``Adam: A method for stochastic optimization,''
  \emph{arXiv preprint arXiv:1412.6980}, 2014.

\bibitem{lecun1998gradient}
Y.~LeCun, L.~Bottou, Y.~Bengio, and P.~Haffner, ``Gradient-based learning
  applied to document recognition,'' \emph{Proc. of the IEEE}, vol.~86, no.~11,
  pp. 2278--2324, 1998.

\bibitem{xiao2017fashion}
H.~Xiao, K.~Rasul, and R.~Vollgraf, ``Fashion-mnist: a novel image dataset for
  benchmarking machine learning algorithms,'' \emph{arXiv preprint
  arXiv:1708.07747}, 2017.

\bibitem{amir2017low}
A.~Amir \emph{et~al.}, ``A low power, fully event-based gesture recognition
  system,'' in \emph{Proc. of the IEEE Conf. on Computer Vision and Pattern
  Recognition}, 2017, pp. 7243--7252.

\bibitem{eshraghian2021training}
J.~K. Eshraghian \emph{et~al.}, ``Training spiking neural networks using
  lessons from deep learning,'' \emph{arXiv preprint arXiv:2109.12894}, 2021.

\bibitem{lu2020exploring}
S.~Lu and A.~Sengupta, ``Exploring the connection between binary and spiking
  neural networks,'' \emph{Frontiers in Neuroscience}, 2020.

\bibitem{o2013real}
P.~O'Connor, D.~Neil, S.-C. Liu, T.~Delbruck, and M.~Pfeiffer, ``Real-time
  classification and sensor fusion with a spiking deep belief network,''
  \emph{Frontiers in Neuroscience}, vol.~7, p. 178, 2013.

\bibitem{neftci2019surrogate}
E.~O. Neftci, H.~Mostafa, and F.~Zenke, ``Surrogate gradient learning in
  spiking neural networks: Bringing the power of gradient-based optimization to
  spiking neural networks,'' \emph{IEEE Signal Process. Mag.}, vol.~36, no.~6,
  pp. 51--63, 2019.

\bibitem{hunsberger2015spiking}
E.~Hunsberger and C.~Eliasmith, ``Spiking deep networks with lif neurons,''
  \emph{arXiv preprint arXiv:1510.08829}, 2015.

\bibitem{hubara2016binarized}
I.~Hubara, M.~Courbariaux, D.~Soudry, R.~El-Yaniv, and Y.~Bengio, ``Binarized
  neural networks,'' \emph{Adv. in Neural Inf. Process. Syst.}, vol.~29, 2016.

\bibitem{bengio2013estimating}
Y.~Bengio, N.~L{\'e}onard, and A.~Courville, ``Estimating or propagating
  gradients through stochastic neurons for conditional computation,''
  \emph{arXiv preprint arXiv:1308.3432}, 2013.

\bibitem{lee2016training}
J.~H. Lee, T.~Delbruck, and M.~Pfeiffer, ``Training deep spiking neural
  networks using backpropagation,'' \emph{Frontiers in Neuroscience}, vol.~10,
  p. 508, 2016.

\bibitem{o2016deep}
P.~O'Connor and M.~Welling, ``Deep spiking networks,'' \emph{arXiv preprint
  arXiv:1602.08323}, 2016.

\bibitem{hinton2012}
G.~Hinton, ``Neural networks for machine learning,'' \emph{Coursera, video
  lectures}, 2012.

\bibitem{putra2021}
R.~V. Putra and M.~Shafique, ``{Q-SpiNN}: A framework for quantizing spiking
  neural networks,'' in \emph{2021 Intl. Joint Conf. on Neural Networks
  (IJCNN)}, 2021.

\bibitem{kheradpisheh2021bs4nn}
S.~R. Kheradpisheh, M.~Mirsadeghi, and T.~Masquelier, ``Bs4nn: Binarized
  spiking neural networks with temporal coding and learning,'' \emph{Neural
  Process. Letters}, pp. 1--19, 2021.

\bibitem{schaefer2020quantizing}
C.~J. Schaefer and S.~Joshi, ``Quantizing spiking neural networks with
  integers,'' in \emph{Intl. Conf. on Neuromorphic Syst. 2020}, 2020, pp. 1--8.

\bibitem{he2016deep}
K.~He, X.~Zhang, S.~Ren, and J.~Sun, ``Deep residual learning for image
  recognition,'' in \emph{Proc. of the IEEE Conf. on computer vision and
  pattern recognition}, 2016, pp. 770--778.

\bibitem{he2019bag}
T.~He, Z.~Zhang, H.~Zhang, Z.~Zhang, J.~Xie, and M.~Li, ``Bag of tricks for
  image classification with convolutional neural networks,'' in \emph{Proc. of
  the IEEE/CVF Conf. on Computer Vision and Pattern Recognition}, 2019, pp.
  558--567.

\bibitem{cordone2021learning}
L.~Cordone, B.~Miramond, and S.~Ferrante, ``Learning from event cameras with
  sparse spiking convolutional neural networks,'' \emph{arXiv preprint
  arXiv:2104.12579}, 2021.

\bibitem{liu2021sstdp}
F.~Liu, W.~Zhao, Y.~Chen, Z.~Wang, T.~Yang, and L.~Jiang, ``Sstdp: Supervised
  spike timing dependent plasticity for efficient spiking neural network
  training,'' \emph{Frontiers in Neuroscience}, vol.~15, 2021.

\bibitem{shen2021backpropagation}
G.~Shen, D.~Zhao, and Y.~Zeng, ``Backpropagation with biologically plausible
  spatio-temporal adjustment for training deep spiking neural networks,''
  \emph{arXiv preprint arXiv:2110.08858}, 2021.

\bibitem{lee2020neuromorphic}
S.-T. Lee and J.-H. Lee, ``Neuromorphic computing using nand flash memory
  architecture with pulse width modulation scheme,'' \emph{Frontiers in
  Neuroscience}, vol.~14, p. 945, 2020.

\bibitem{akiba2019optuna}
T.~Akiba, S.~Sano, T.~Yanase, T.~Ohta, and M.~Koyama, ``Optuna: A
  next-generation hyperparameter optimization framework,'' in \emph{Proc. of
  the 25th ACM SIGKDD Intl. Conf. on knowledge discovery \& data mining}, 2019,
  pp. 2623--2631.

\bibitem{bergstra2011algorithms}
J.~Bergstra, R.~Bardenet, Y.~Bengio, and B.~K{\'e}gl, ``Algorithms for
  hyper-parameter optimization,'' \emph{Adv. in Neural Inf. Process. Syst.},
  vol.~24, 2011.

\bibitem{shrestha2018slayer}
S.~B. Shrestha and G.~Orchard, ``Slayer: spike layer error reassignment in
  time,'' in \emph{Proc. of the 32nd Intl. Conf. on Neural Inf. Process.
  Syst.}, 2018, pp. 1419--1428.

\bibitem{perez2021neural}
N.~Perez-Nieves, V.~C. Leung, P.~L. Dragotti, and D.~F. Goodman, ``Neural
  heterogeneity promotes robust learning,'' \emph{bioRxiv}, pp. 2020--12, 2021.

\bibitem{brevitas}
\BIBentryALTinterwordspacing
A.~Pappalardo, ``Xilinx/brevitas,'' \emph{Zenodo}, 2021. [Online]. Available:
  \url{https://doi.org/10.5281/zenodo.3333552}
\BIBentrySTDinterwordspacing

\bibitem{paszke2019pytorch}
A.~Paszke \emph{et~al.}, ``Pytorch: An imperative style, high-performance deep
  learning library,'' \emph{Adv. in Neural Inf. Process. Syst.}, vol.~32, pp.
  8026--8037, 2019.

\bibitem{patrick2008128x}
L.~Patrick, C.~Posch, and T.~Delbruck, ``A 128x 128 120 db 15$\mu$ s latency
  asynchronous temporal contrast vision sensor,'' \emph{IEEE J. of Solid-State
  Circuits}, vol.~43, pp. 566--576, 2008.

\bibitem{sharmin2019comprehensive}
S.~Sharmin, P.~Panda, S.~S. Sarwar, C.~Lee, W.~Ponghiran, and K.~Roy, ``A
  comprehensive analysis on adversarial robustness of spiking neural
  networks,'' in \emph{2019 International Joint Conference on Neural Networks
  (IJCNN)}.\hskip 1em plus 0.5em minus 0.4em\relax IEEE, 2019, pp. 1--8.

\bibitem{yousefzadeh2019conversion}
A.~Yousefzadeh, S.~Hosseini, P.~Holanda, S.~Leroux, T.~Werner,
  T.~Serrano-Gotarredona, B.~L. Barranco, B.~Dhoedt, and P.~Simoens,
  ``Conversion of synchronous artificial neural network to asynchronous spiking
  neural network using sigma-delta quantization,'' in \emph{2019 IEEE
  International Conference on Artificial Intelligence Circuits and Systems
  (AICAS)}.\hskip 1em plus 0.5em minus 0.4em\relax IEEE, 2019, pp. 81--85.

\end{thebibliography}
\end{document}